\documentclass[10pt,twocolumn,letterpaper]{article}

\usepackage[pagenumbers]{cvpr} 

\usepackage{graphicx}
\usepackage{amsmath}
\usepackage{amssymb}
\usepackage{booktabs}
\usepackage{times}
\usepackage{epsfig}
\usepackage{graphicx}
\usepackage{amsmath}
\usepackage{amssymb}
\usepackage{url}
\def\blu#1{\textbf{\color{blue} #1}} 
\def\red#1{\textbf{\color{red}  #1}} 
\usepackage{tabularx}
\usepackage{multirow}

\usepackage[pagebackref,breaklinks,colorlinks]{hyperref}

\usepackage[capitalize]{cleveref}
\crefname{section}{Sec.}{Secs.}
\Crefname{section}{Section}{Sections}
\Crefname{table}{Table}{Tables}
\crefname{table}{Tab.}{Tabs.}

\def\cvprPaperID{*****} 
\def\confName{CVPR}
\def\confYear{2022}

\begin{document}

\title{ Reflection Invariance Learning for Few-shot Semantic Segmentation}
\author{
Qinglong Cao$^{1,2}$
\quad
Yuntian Chen$^{2}$
\quad
Chao Ma$^{1}$
\quad
Xiaokang Yang$^{1}$\\
$^1$ MoE Key Lab of Artificial Intelligence, AI Institute, Shanghai Jiao Tong University, Shanghai, China\\
$^2$ Eastern Institute for Advanced Study, Zhengjiang, China\\
{\tt \{caoql2022, chaoma, xkyang\}@sjtu.edu.cn}\\
{\tt ychen@eias.ac.cn}
}



\maketitle

\begin{abstract}

Few-shot semantic segmentation (FSS) aims to segment objects of unseen classes in query images with only a few annotated support images. Existing FSS algorithms typically focus on mining category representations from the single-view support to match semantic objects of the single-view query. However, the limited annotated samples render the single-view matching struggle to perceive the reflection invariance of novel objects, which results in a restricted learning space for novel categories and further induces a biased segmentation with demoted parsing performance. To address this challenge, this paper proposes a fresh few-shot segmentation framework to mine the reflection invariance in a multi-view  matching manner.  Specifically, original and reflection support features from different perspectives with the same semantics are learnable fused to obtain the reflection invariance prototype with a stronger category representation ability. Simultaneously, aiming at providing better prior guidance, the Reflection Invariance Prior Mask Generation (RIPMG) module is proposed to integrate prior knowledge from different perspectives. Finally, segmentation predictions from varying views are complementarily merged in the Reflection Invariance Semantic Prediction (RISP) module to  yield precise segmentation predictions.  Extensive experiments on both PASCAL-$5^\textit{i}$  and COCO-$20^\textit{i}$ datasets demonstrate the effectiveness of our approach and show that our method could achieve state-of-the-art performance. Code is available at \url{https://anonymous.4open.science/r/RILFS-A4D1}

\end{abstract}

\section{Introduction}
With the development of deep learning technology, semantic segmentation task ~\cite{FCN,segnet,deeplab,PSP,mask,atrous,u-net} has achieved promising progress. Most advanced semantic segmentation algorithms follow the fully-supervised learning manner and require large-scale annotated data for training. However, acquiring quantities of annotated data is very time-consuming and laborious. To tackle this realistic dilemma, few-shot semantic segmentation (FSS) task has been proposed to efficiently parse the unseen-category objects with only a few annotated support samples. 

\begin{figure}[t]
	\begin{center}
		\includegraphics[width=1.0\linewidth]{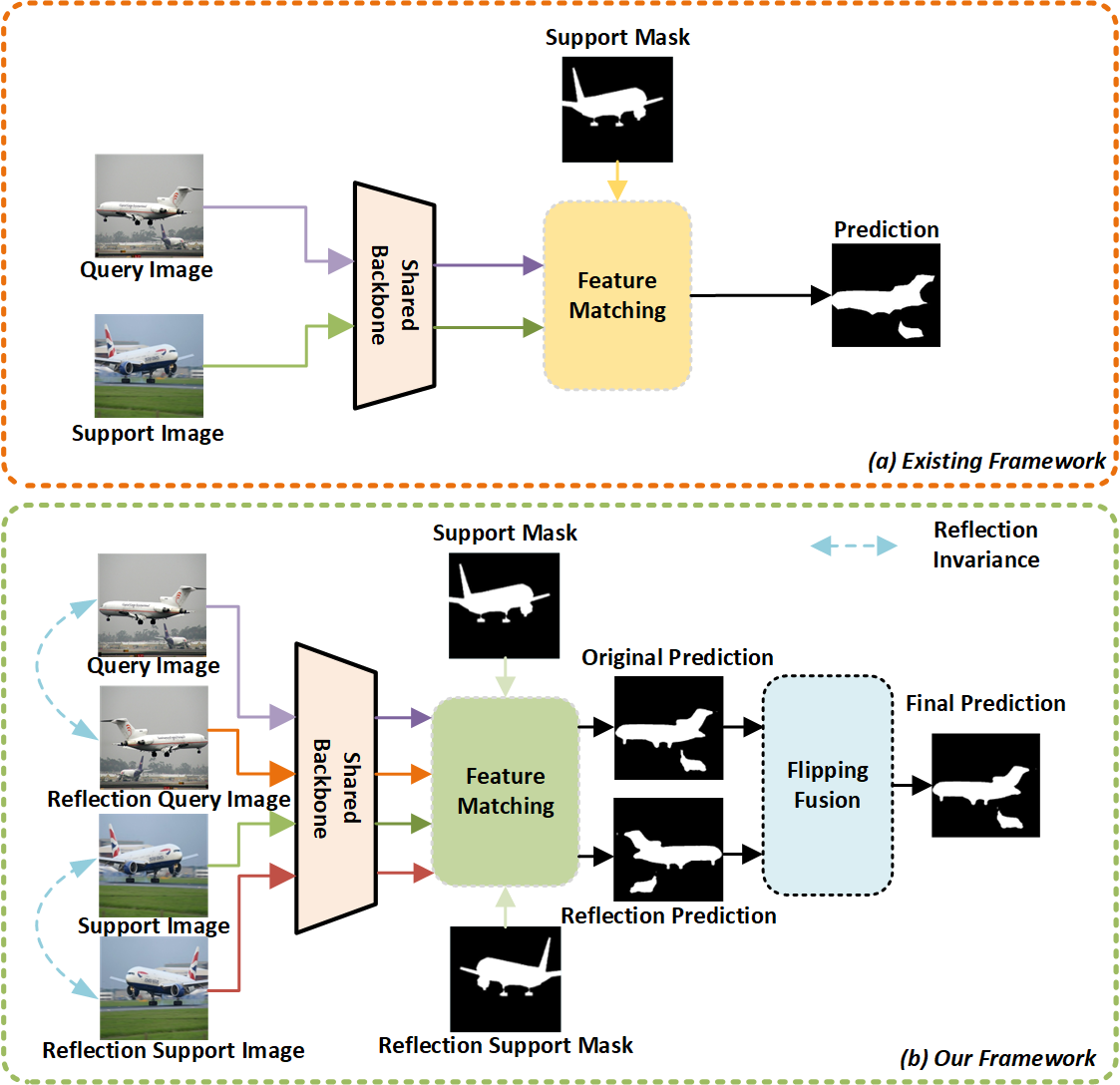}
	\end{center}
	\caption{The comparison between the existing framework and our framework for few-shot semantic segmentation.     }
	\label{fig:first}
 \vspace{-0.6cm}
\end{figure}

Recently, Few-shot semantic segmentation task~\cite{johnander2022dense,PMMs,HPA,beyond_prototype,fan2022self,liu2022learning,liu2023fecanet,cheng2022holistic,lang2022learning} has obtained increasing attention and many researchers have proposed diverse sophisticated FSS algorithms. The early FSS algorithms~\cite{OSLSM,panet,canet} first utilized the mask average pooling on the support features to generate support prototypes as the category representations.
Then, by directly performing a dense matching between support prototypes and query features, the category-related semantic objects could be successfully parsed. Following this pipeline, some works~\cite{PMMs,HPA,beyond_prototype,liu2020part} attempted to provide better support prototypes with part decomposition. Meanwhile, some researchers~\cite{xie2021scale, min2021hypercorrelation,PFEnet} utilized multi-scale or multi-layer matching to provide more precise segmentation predictions. Moreover, working on how to eliminate distracting objects in the parsing process, some FSS methods~\cite{liu2022learning,lang2022learning} have achieved great performance.  

\begin{figure}[t]
	\begin{center}
		\includegraphics[width=1.0\linewidth]{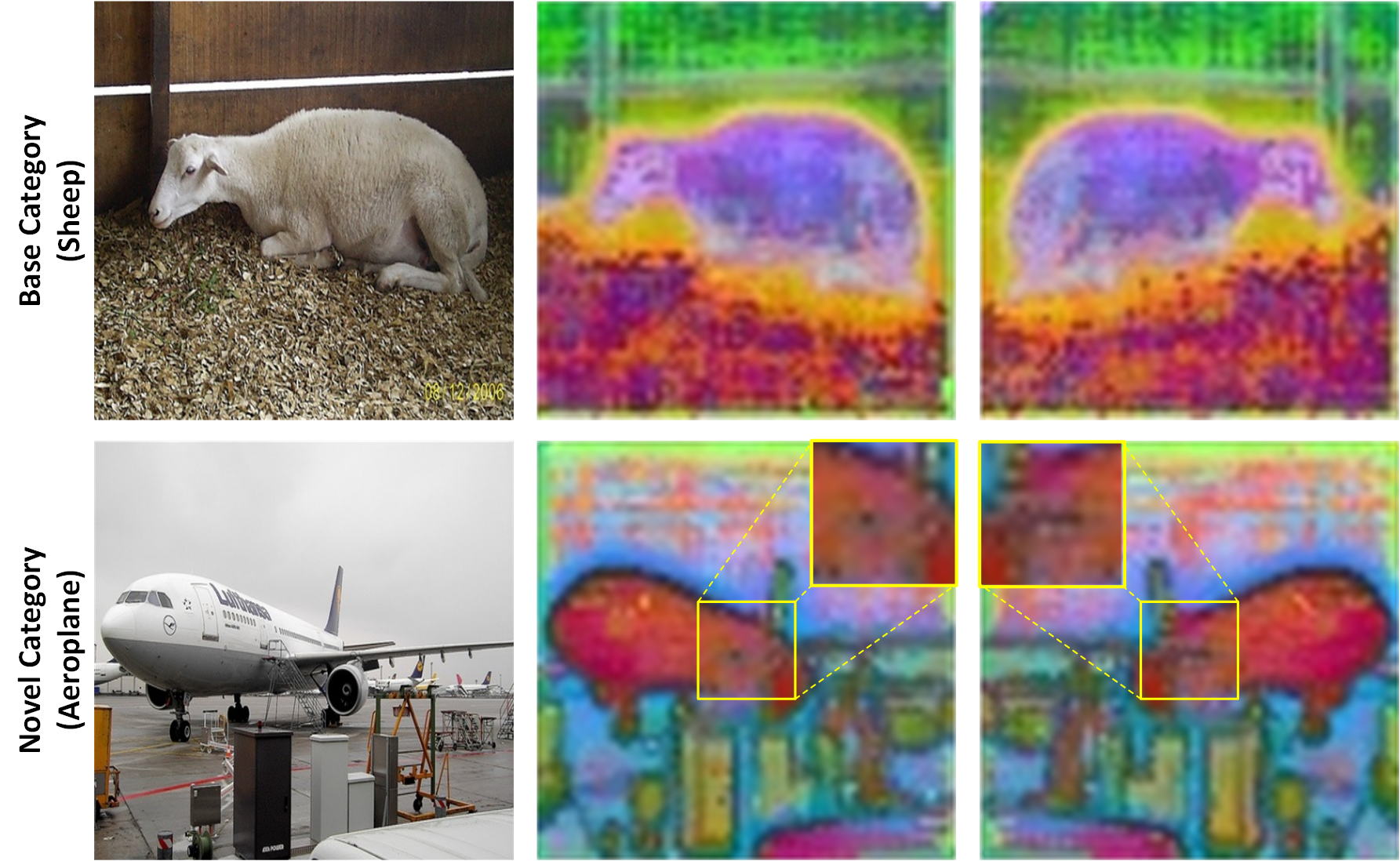}
	\end{center}
	\caption{Multi-view visualizations of  features from both base and novel categories.      }
	\label{insight}
 \vspace{-0.5cm}
\end{figure}
Though previous FSS algorithms have obtained some achievements, they only focus on how to precisely parse semantic objects of query images in a single-view matching manner~(Figure \ref{fig:first} (a)). However, as FSS networks are typically meta-trained using abundant annotated base categories, they tend to exhibit weak reflection invariance perception ability when presented with limited labeled single-view novel objects. For instance, as illustrated in Figure \ref{insight}, with Fold-0 in PASCAL-$5^\textit{i}$ ~\cite{OSLSM} and BAM method ~\cite{lang2022learning}, the sheep from  base categories exhibit similar features across different views, while the novel category aeroplane shows obvious differences between multi-view features. 
Thus, existing single-view matching paradigms would result in a constrained and incomplete learning space for the novel categories, which further leads to biased segmentation results and unsatisfactory accuracy.
To handle this problem,  we propose a novel multi-view matching FSS method to mine the semantic reflection invariance~( Figure \ref{fig:first} (b)). 

Specifically,  for the support images,  the  extracted original and reflection support features are both going through the mask average pooling to obtain the original  and reflection prototypes. Then, Through learnable parameters, the original prototypes and reflection prototypes are fused to generate reflection invariance prototypes with stronger category representative ability. For the query images,  original and reflection query features are respectively propagated into the reflection Invariance Prior Mask Generation (RIPMG) module to acquire the corresponding reflection invariance prior masks. Subsequently, under the guidance of reflection invariance prior prototypes and reflection invariance masks, the original and  reflection query features are inputted into a segmentation network to achieve the corresponding segmentation predictions. Finally, the  predictions from different views are learnable fused in the Reflection Invariance  Semantic Prediction (RISP) module to obtain the final predictions. Additionally, considering the  scene differences between query images and   support images,  the adjustment factor  ~\cite{lang2022learning} is also applied to both original and reflection query images.

To sum up, the main contributions of our proposed method could be concluded as follows:
\begin{itemize}
	\item To the best of our knowledge, this is the first time reflection invariance is explored in the FSS task with multi-view matching, which could help the network more efficiently utilize the limited annotated novel class samples. 
	
	\item Through mining support reflection invariance, the generated reflection invariance prototypes provide better support guidance. Also, with the help of RIPMG and RISP, the proposed network successfully fuses the predictions from different perspectives to acquire better segmentation results.
	
	\item Experimental results on PASCAL-$5^\textit{i}$  and COCO-$20^\textit{i}$  datasets demonstrate that the proposed method achieves state-of-the-art performance across 1-shot and 5-shot settings.
	
\end{itemize}

\section{Related Work}
\textbf{Semantic Segmentation.} Semantic segmentation aims at precisely classifying each pixel of images.  However, with the rising of deep learning technology and convolutional neural networks(CNNs)~\cite{imagenet}, great progress~\cite{FCN,u-net,PSP,refinenet,atrous,zhong2020squeeze,fu2019dual,huang2019ccnet} has been produced in the semantic segmentation domain. Particularly, FCN ~\cite{FCN} firstly performs precise pixel-prediction with fully convolutional layers. Based on FCN, various advanced networks have been proposed. For instance, U-net ~\cite{u-net} was proposed to segment the medical image in an encoder-decoder architecture. Meanwhile, some researchers leveraged multi-scale image features~\cite{PSP,refinenet} to boost the segmentation performance. Focusing on refining the featuring extracting ability, atrous convolution~\cite{atrous} was introduced to handle the semantic segmentation task. Additionally,   inspired by the efficient attention mechanism, some researchers utilized the attention-based network~\cite{zhong2020squeeze,fu2019dual,huang2019ccnet} to enhance the parsing performance. Recently, since the transformer architecture has shown powerful feature extracting and processing ability,  some transformer-based methods~\cite{xie2021segformer,strudel2021segmenter} have been proposed to handle the semantic segmentation task and achieved new milestones. 

\textbf{Few-shot Semantic Segmentation.} Acquiring high-quality annotated training data is very time-consuming and laborious. Thus, it is very urgent to develop few-shot semantic segmentation algorithms that parse the unseen-category objects with only a few labeled samples. Typically, support prototypes ~\cite{dong2018few} would be first extracted from the annotated support images to construct the category representations. By computing similarities between prototypes and pixels of query images, semantic objects of the target category could be easily located. Based on this pipeline, plenties of great algorithms for few-shot semantic segmentation~\cite{dong2018few,canet,panet,beyond_prototype,shi2022dense,PMMs,liu2022learning,lang2022learning,HPA,fan2022self,liu2022intermediate} have been conducted. For instance, through an iterative optimization process, CAnet ~\cite{canet} efficiently refined the prediction results in an iterative learning manner. Moreover, PAnet~\cite{panet} simultaneously leveraged the query-support matching and support-query matching as dual supervision to directly boost the segmentation performance. Aiming at mining better relations between query and support, Shi \textit{et al.} ~\cite{shi2022dense} designed a dense cross-query-and-support attention to enhance the prediction results. Focusing on providing better prototypes, DCP~\cite{beyond_prototype} and PMMs~\cite{PMMs} divided the support features into diverse proxies and prototypes representing different semantic regions, which clearly could provide better support guidance for query images. Working on eliminating the distracting objects, NERTNet ~\cite{liu2022learning} designed an efficient network to learn the non-target knowledge. Simultaneously,  by introducing another branch to remove the biased base knowledge, BAM ~\cite{lang2022learning} brought a new perspective for few-shot semantic segmentation.  To alleviate the intra-class gap between query and support, SSP ~\cite{fan2022self} successfully boosted segmentation performance in a self-support learning manner.  Furthermore, IPMT ~\cite{liu2022intermediate} refines the query-support matching process through mining the intermediate prototype. Working on multi-scale feature matching, HSNet ~\cite{min2021hypercorrelation} achieves new improvements with 4D convolutional layers.  

Though existing FSS algorithms have achieved great progress, these methods only focus on how to precisely segment semantic objects of query in a single-view matching manner. The inner  semantic reflection invariance of both query and support images has rarely been explored, which could clearly result in the ineffective usage of annotated training samples.

\textbf{Invariance Learning.} Invariance is a valid and worth-exploring physical property
for many domains.  For instance, Sun \textit{et al.} ~\cite{sun2022probabilistic} introduce the probabilistic invariance into the multi-agent dynamics for improved trajectory forecasting. Moreover, aiming at accelerating the optimization process,  Hao \textit{et al.} ~\cite{zhao2022symmetry} derived a loss-invariant group to implement the symmetry teleportation. To automatically discover the invariance, Dehmamy \textit{et al.} ~\cite{dehmamy2021automatic} developed the lie algebra convolutional network with the direct connections between L-conv and physics.  Objects in remote sensing images are often distributed with arbitrary orientation. Naturally, the objects with random orientations have the same semantics as the original objects. Based on this invariance, Cheng \textit{et al.} ~\cite{cheng2016learning} and  Han \textit{et al.} ~\cite{han2021redet}proposed a rotation-invariant convolutional neural
network to efficiently perform object detection in remote sensing images.

 Inspired by the semantic reflection invariance in natural images, this paper first introduces semantic reflection invariance into few-shot semantic segmentation and achieves great performance.

 \begin{figure*}[t]
	\begin{center}
		\includegraphics[width=1.0\linewidth]{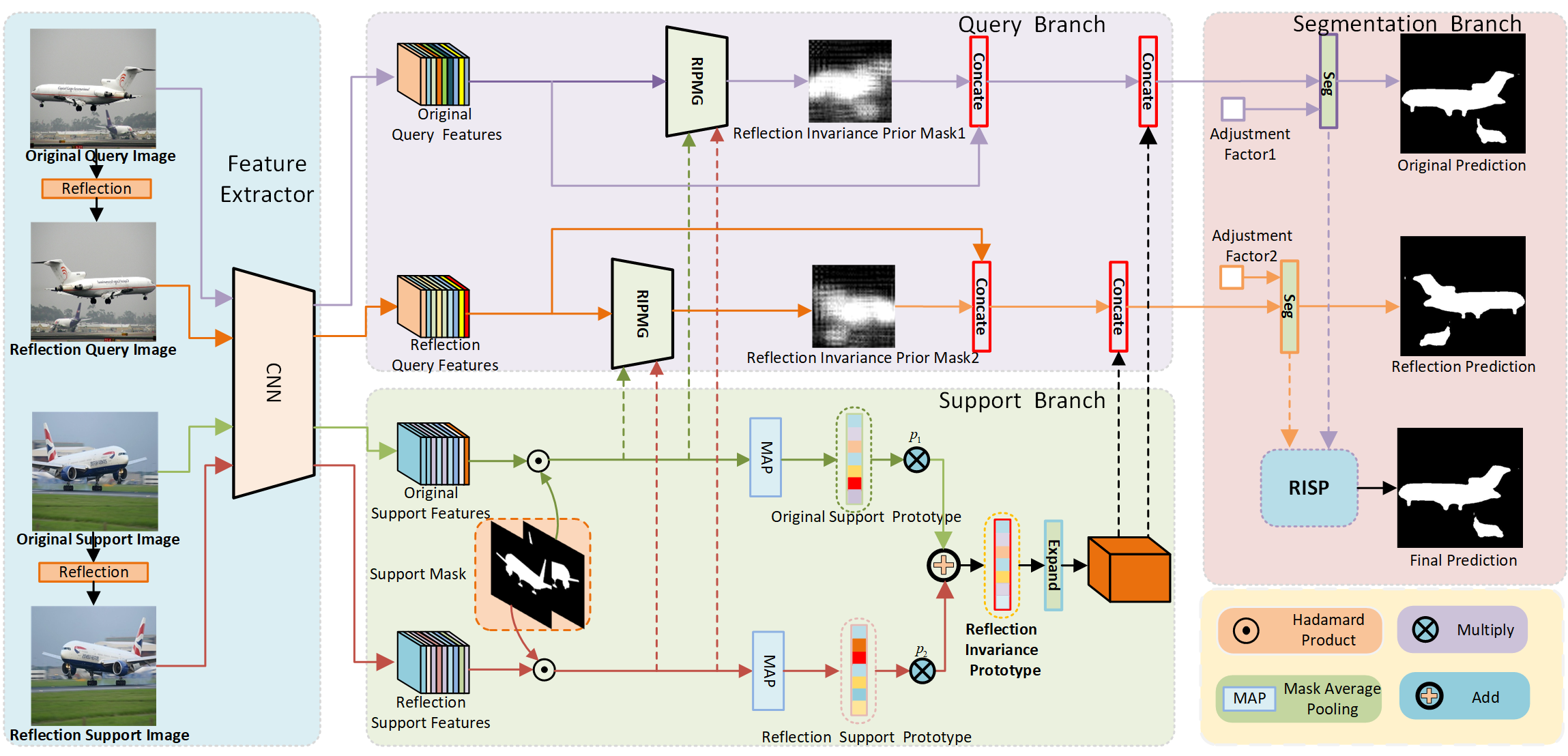}
	\end{center}
	\caption{Overall architecture of our proposed network. Solid/dotted lines are utilized to denote horizontal/vertical direction for better visualization. In the support branch, the original support features and reflection support features are fused to generate the reflection invariance prototypes with stronger category representative ability. For the query branch, under the guidance of  support reflection invariance information, query features from different perspectives are propagated into the segmentation network to provide different predictions, which are further fused to provide the reflection invariance predictions.   }
	\label{fig:second}
  \vspace{-0.4cm}
\end{figure*}

\section{Proposed Method}

\subsection{Problem Definition}
In a few-shot semantic segmentation task, the categories of the whole dataset would be divided into two non-overlapping subsets of classes, namely $C_{base}$ and $C_{novel}$. The images containing objects from $C_{base}$  constitute the training dataset  $D_{train}$. Correspondingly, the testing dataset $D_{test}$ is the accumulation of images containing objects from $C_{novel}$.  The goal of FSS is to learn the meta knowledge from  abundant annotated $C_{base}$ and transfer the meta knowledge to the $C_{novel}$ with only a few annotated samples. Thus, the few-shot semantic segmentation follows the meta-learning paradigm and leverages the episodic training strategy to perform the training process. Particularly, in $K$-shot setting scenario,  $K+1$ labeled images $\left\{ (I_s^1,M_s^1),(I_s^2,M_s^2), \cdot  \cdot  \cdot (I_s^k,M_s^k),({I_q},{M_q})\right\}$ of targeted category would be sampled from the  $D_{train}$ episodically, where $(I_s^i,M_s^i)$  denotes the support image-mask pair and $({I_q},{M_q})$ denotes the query image-mask pair. During the training phase, the goal of the  model is to make precise prediction  ${\hat M_q}$ of ${I_q}$ under the guidance of the $K$-shot support image-mask pairs, and the model parameters are iteratively updated.
After the training process, the model would be directly evaluated on the testing dataset $D_{test}$ to show segmentation performance on novel classes.

\subsection{Method Overview}
The overall model architecture is shown in Figure \ref{fig:second}. 
First, with the horizontal flip operation, we obtain the reflection images of both query and support images. Then, a shared feature extractor is leveraged to extract the hierarchical features of these images. After that, in the support branch, through the mask average pooing (MAP) on original and reflection support features, the network would obtain original  and reflection support prototypes. By learnable fusing original and reflection support prototypes, the reflection invariance prototype could be acquired. In the query branch, the proposed Reflection Invariance Prior Mask Generation (\textbf{RIPMG}, Section \ref{RIPMG} ) module could respectively provide reflection invariance prior masks for both original and reflection query features.  Subsequently, utilizing the reflection invariance prior masks, reflection invariance prototypes, and the adjustment factors ~\cite{lang2022learning} as support guidance, the segmentation network ~\cite{liu2022learning} would perform segmentation for both original and reflection query features. Finally, through the Reflection Invariance Semantic Prediction (\textbf{RISP}, Section \ref{RISP} ) module, the original and reflection predictions are fused to acquire more precise reflection invariance  predictions.  

\subsection{Reflection Invariance Prototype Creation} \label{FIPC}
Given the support image $I_s \in \mathbb{R}^{3 \times H \times W} $ and query image  $I_q \in \mathbb{R}^{3 \times H \times W} $, the original support and query images are firstly horizontally flipped to obtain the reflection support images $I_s^h \in \mathbb{R}^{3 \times H \times W} $ and reflection query images $I_q^h \in \mathbb{R}^{3 \times H \times W} $. Then, by propagating these images into the pre-trained feature extractor (i.e., Resnet50 ~\cite{he2016deep}, VGG16~\cite{simonyan2014very}),  original query features  $F_q \in \mathbb{R}^{C \times H \times W} $, original support features  $F_s \in \mathbb{R}^{C \times H \times W} $, reflection query features $F_q^h \in \mathbb{R}^{C \times H \times W} $, and reflection support features $F_s^h \in \mathbb{R}^{C \times H \times W} $ are generated. 

The joint utilization of  original support features $F_s$ and reflection support features $F_s^h$ from different perspectives with the same semantics could provide better support information for support images. Firstly, the mask average pooling operation is applied to the original support features $F_s$ and reflection support features $F_s^h$ to acquire original support prototype $P_o$  and reflection support prototype $P_f$. To better fuse the two generated prototypes, we set two learnable parameters $p_1$ and $p_2$, which are directly leveraged to fuse the two prototypes in a weighted sum manner. The fusion process could be computed as follow:
\begin{equation}
{P_{FI}} = {p_1} * {P_o} + {p_2}*{P_f}
\end{equation}
where the ${P_{FI}}$ denotes the reflection invariance prototype.

\subsection{Reflection Invariance Prior Mask Generation} \label{RIPMG}

\begin{figure}[t]
	\begin{center}
		\includegraphics[width=1.0\linewidth]{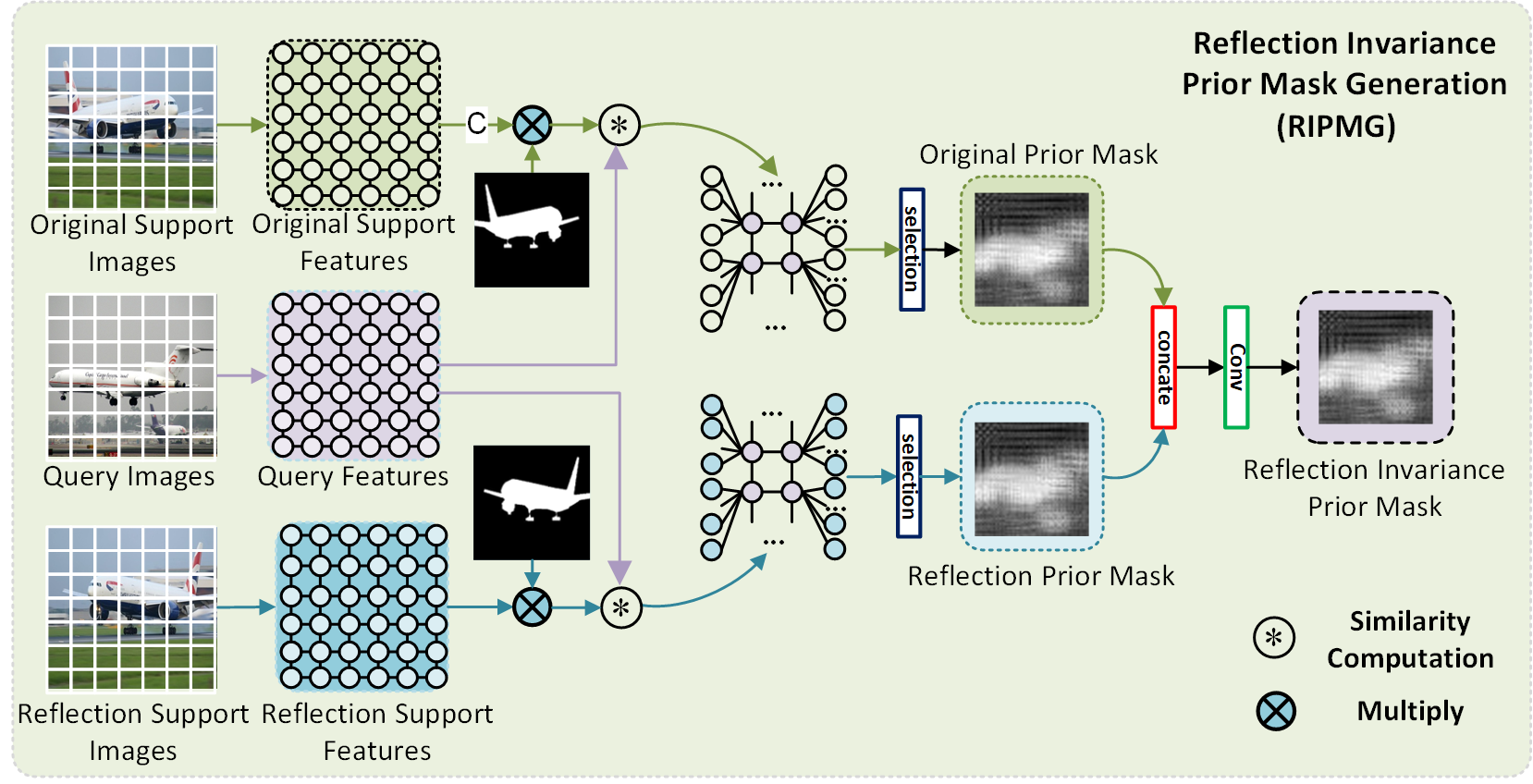}
	\end{center}
	\caption{The illustration of the proposed reflection invariance prior mask generation module. The query images simultaneously denote the original and reflection query images.   }
	\label{fig:third}
  \vspace{-0.4cm}
\end{figure}

To provide better support prior guidance for the segmentation of query images, the reflection invariance prior mask is generated for both $F_q$ and $F_q^h$. The details of the generation process are shown in Figure \ref{fig:third}, where the query images simultaneously denote the original query images and reflection query images. Here, we set the query images as  original query images $I_q$ for a more detailed explanation. First, original support features $F_s$, reflection support features $F_s^h$, and original query features $F_q$ are respectively flattened in spatial dimension as original support features set$\left\{ {f_1^s, f_2^s, \cdot  \cdot  \cdot ,f_N^s} \right\}$, reflection support features set $\left\{ {f_1^{sh}, f_2^{sh}, \cdot  \cdot  \cdot ,f_N^{sh}} \right\}$ and query features set $\left\{ { f_1^q,f_2^q, \cdot  \cdot  \cdot , f_N^q} \right\}$. $f \in \mathbb{R}^C$ and $N= H \times W$. Then, to filter out the valid support information, the support mask $M_s$ and reflection support mask $M_s^h$ are leveraged to obtain the foreground features:

\begin{equation}
\left\{ { f_1^s,f_2^s, \cdot  \cdot  \cdot , f_i^s} \right\} = {F_s} \odot {M_s}
\end{equation}

\begin{equation}
\left\{ { f_1^{sh},f_2^{sh}, \cdot  \cdot  \cdot , f_j^{sh}} \right\} = {F_s^h}  \odot {M_s^h}
\end{equation}
where $\odot$ denotes the Hadamard product and $i,j$ denote the index of foreground.
Then, the cosine similarity  is leveraged as the correlation function to compute the pixel-level relation score between query features and original/reflection support features:
\begin{equation}
\cos (f_n^q,f_i^s) = \frac{{f_n^q f_i^s}}{{\left\| {f_n^q} \right\|\left\| {f_i^s} \right\|}},n \in \{ 1,2,....,N\} 
\end{equation}

\begin{equation}
\cos (f_n^q,f_j^{sh}) = \frac{{f_n^q f_j^{sh}}}{{\left\| {f_n^q} \right\|\left\| {f_j^{sh}} \right\|}},n \in \{ 1,2,....,N\} 
\end{equation}

For each element of the query features set, the highest relation score among all support elements is selected as the prior value:
\begin{equation}
s_n^s = \mathop {\max }\limits_{i \in foreground } \cos (f_n^q,f_i^{s})
\end{equation}
\begin{equation}
s_n^{sh} = \mathop {\max }\limits_{j \in foreground} \cos (f_n^q,f_j^{sh})
\end{equation}
where $s_n^s$ denotes the $n-th$ element of original prior mask and $s_n^{sh}$ denotes the $n-th$ element of reflection prior mask. In this way, the max probability of each element belonging to the foreground has been computed. By accumulating these scores, we could obtain the original prior mask $m_o^p = \left\{ { s_1^s, s_2^s, \cdot  \cdot  \cdot , s_N^s} \right\}$ and reflection prior mask $m_f^p = \left\{ { s_1^{sh}, s_2^{sh}, \cdot  \cdot  \cdot , s_N^{sh}} \right\}$.
To efficiently provide support information for query images with reflection invariance prior mask $m_{FI}^p$, the concatenation operation and a $1 \times 1$ convolutional layer $\mathcal{F}$ is leveraged to fuse the original prior mask and reflection prior mask:
\begin{equation}
{m_{FI}^p} = \mathcal{F}(concate({m_f^p},{m_o^p}))
\end{equation}
By performing the above operation for both original query images and reflection query images, the network provides reflection invariance prior masks for the two query images.

\subsection{Reflection Invariance Semantic Prediction} \label{RISP}
\begin{figure}[t]
	\begin{center}
		\includegraphics[width=1.0\linewidth]{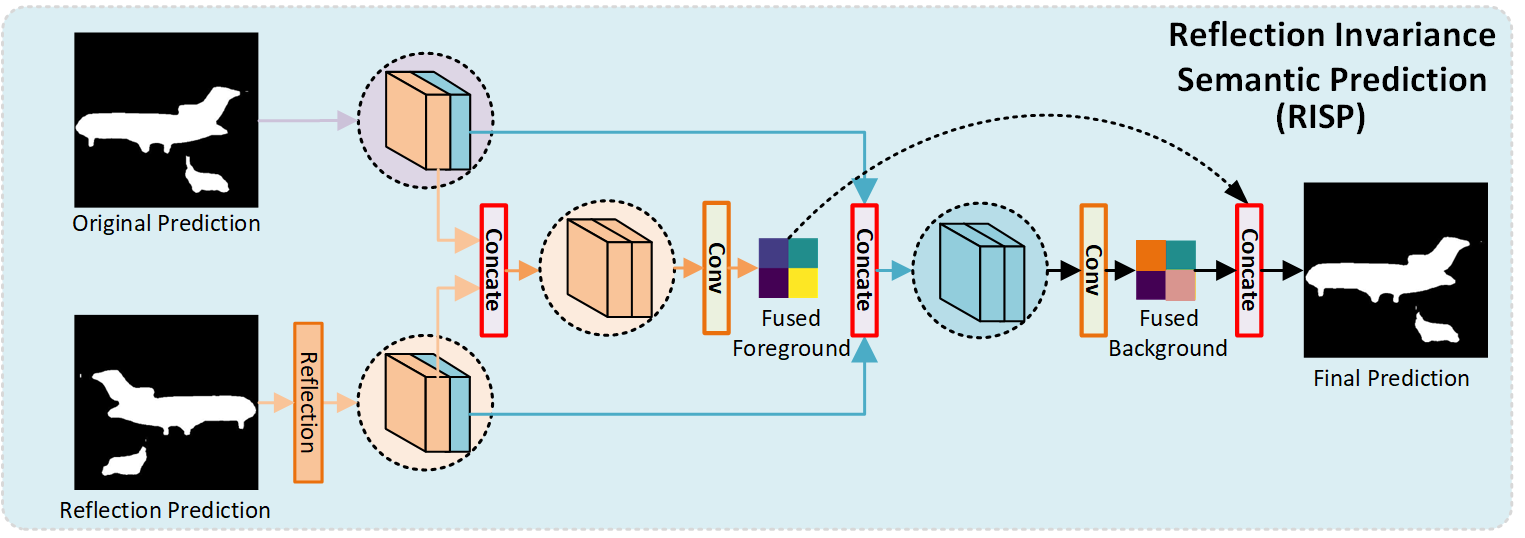}
	\end{center}
	\caption{The illustration of the proposed reflection invariance semantic prediction module. By fusing the predictions from different perspectives, the finally generated reflection invariance  predictions would have higher accuracy.  }
	\label{fig:fourth}
  \vspace{-0.5cm}
\end{figure}

Different perspectives of images would result in varying segmentation results, thus we could fuse the original and reflection segmentation results to obtain more precise reflection invariance  predictions. Particularly, the reflection invariance prototype $P_{FI}$ is expanded as
${\widehat P_{FI}} \in \mathbb{R}^{C \times H \times W}$, which has the same size as original/reflection query features.  Following the procedure in 
 ~\cite{lang2022learning}, ${\widehat P_{FI}}$ and the corresponding reflection invariance prior mask ${m_{FI}^p}$ are jointly leveraged in the segmentation network $\mathcal{F}_{seg}$ ~\cite{liu2022learning} to activate the original query features $F_q$ and the reflection query features $F_q^H$:
 \begin{equation}
{\widehat F_q} = \mathcal{F}_{seg}(concate(F_q,{\widehat P_{FI}},{m_{FI}^{p1}}))
\end{equation}
 \begin{equation}
{\widehat F_q^h} = \mathcal{F}_{seg}(concate(F_q^h,{\widehat P_{FI}},{m_{FI}^{p2}}))
\end{equation}
where $p1$ and $p2$ respectively mean the reflection invariance prior mask for original and reflection query images.
Subsequently, to better boost the efficiency of segmentation, we adapt the adjustment factor to alleviate the scene differences between query images and support images. However, since the network simultaneously leverages original and reflection support images to provide support guidance, we set two learnable parameters $a_1$ and $a_2$ to fuse the adjustment factors ($\theta_o,\theta_h $) from original support images and reflection support images:
 \begin{equation}
{\theta_f} = {a_1} * {\theta_o} + {a_2}*{\theta_h}
\end{equation}
It is noticeable that the network provides different fused adjustment factors(${\theta_f^1, \theta_f^2}$) for original query images and reflection query images. Additionally, the base learner ~\cite{lang2022learning} is also adopted in our network to eliminate the biased base knowledge. Naturally, the network would generate the original prediction $R_o$ and reflection prediction $R_f$. As shown in Figure \ref{fig:fourth},  foreground $R_o^1$ of the original prediction $R_o$ and  foreground $R_f^1$ of reflection prediction $R_f$ are firstly concatenated and fused through a $1 \times 1$ convolutional layer $\mathcal{F}_{fuse}$ to obtain the fused foreground $R_{ff}$:
 \begin{equation}
{R_{ff}^1} = \mathcal{F}_{fuse}(concate(R_o^1,R_f^1))
\end{equation}
Similarly, the fused background $R_{fb}$ is acquired through the concatenation and convolutional fusion of  background $R_o^0$ of the original prediction $R_o$ and  foreground $R_f^0$ of reflection prediction $R_f$:
 \begin{equation}
{R_{ff}^0} = \mathcal{F}_{fuse}(concate(R_o^0,R_f^0))
\end{equation}
By directly concatenating the fused foreground and the fused background, we could obtain the final predictions $R_{ff}$, namely the reflection invariance predictions.
 
\subsection{Training Loss}
To directly supervise the original prediction, reflection prediction, and final prediction, the training phase should have three different binary cross-entropy losses:
\begin{equation}
\begin{split}
    {L_{final}} & =BCE(R_{ff},M_{GT}) \\ & + \alpha BCE(R_o,M_{GT}) +\beta BCE(R_f,M_{GT}^H) 
\end{split}
\end{equation}
where $BCE$ means  binary cross-entropy loss, $M_{GT}$ denotes the ground truth mask, and  $M_{GT}^H$ illustrates the reflection ground truth mask. We set $\alpha$ and $\beta$ as 0.5 for all experiments.
Apart from these supervisions, since we follow the procedure in ~\cite{lang2022learning} to simultaneously leverage the base and meta-learner to eliminate the biased base knowledge, the meta loss $L_{meta}$ is also needed to supervise the initial predictions of original prediction, reflection prediction. Therefore, the total loss $L_{all}$ during the meta-training phase could be defined as follow:
\begin{equation}
    {L_{all}} = L_{final} +  L_{meta}
\end{equation}


\begin{table*}[t!]
	\centering
	\scriptsize
	\footnotesize
	\renewcommand{\arraystretch}{0.9}
	\renewcommand{\tabcolsep}{2.8mm}		
	\caption{  mIoU performance on four folds of PASCAL-$5^\textit{i}$. \red{Red} and \blu{blue} are leveraged to indicate the two best scores.}
	\scalebox{0.80}{
		\begin{tabular}{cc|cccccc|cccccc}
			\hline
			\multicolumn{2}{c|}{\multirow{2}{*}{Method}}&  \multicolumn{6}{c|}{1-shot}&\multicolumn{6}{c}{5-shot} \\
			
			\multicolumn{2}{c|}{}     
			& Fold-0  & Fold-1 & Fold-2 & Fold-3 & Mean &FB-IoU & Fold-0  & Fold-1 & Fold-2 & Fold-3 & Mean &FB-IoU\\ 
			\hline \hline
			\multicolumn{14}{c}{VGG16-Backbone}  \\ \hline  
			\multicolumn{2}{c|}{OSLSM (TCYB'19) \cite{OSLSM}}   & 33.60 & 55.30 & 40.90 & 33.50 & 40.80 &61.30 & 35.90 & 58.10 & 42.70 & 39.10 & 44.00 &61.50  \\ 
			\multicolumn{2}{c|}{PANet (ICCV'2019) \cite{panet}}   & 42.30 & 58.00 & 51.10 & 41.20 & 48.10 &66.50 & 51.80 & 64.60 & 59.80 & 46.50 & 55.70 &70.70\\ 
			\multicolumn{2}{c|}{PFENet (TPAMI'2020) \cite{PFEnet}}    & 56.90 & 68.20 & 54.40 & 52.40 & 58.00 &72.00 & 59.00 & 69.10 & 54.80 & 52.90 & 59.00 &72.30\\ 
			\multicolumn{2}{c|}{HSNet (ICCV'2021) \cite{ min2021hypercorrelation}} &59.60 &65.70 &59.60 &54.00 &59.70 &73.40   &64.90 &69.00 &64.10 &58.60 &64.10 &76.60\\ 
   			\multicolumn{2}{c|}{NTRENet (CVPR'2022) \cite{liu2022learning}} &57.70 &67.60 &57.10 &53.70 &59.00 &73.10 &60.30 &68.00 &55.20 &57.10 &60.20 &74.20\\ 
   			\multicolumn{2}{c|}{DCP (IJCAI'2022) \cite{beyond_prototype}}&59.67 &68.67 &63.78 &53.11 &61.31 &74.90 &64.25 &70.66 &67.38 &61.08 &65.84 & 79.4\\ 
                \multicolumn{2}{c|}{DACM (ECCV'2022) \cite{xiong2022doubly}} &61.80 &67.80 &61.40 &56.30 &61.80 &75.50 &66.10 &70.60 &65.80 &60.20 &65.70 &77.80\\
                \multicolumn{2}{c|}{HPA (TPAMI'2022) \cite{HPA}} &61.40 &69.81 &63.84 &54.13 &62.30 &75.20 &65.42 &71.74 &68.80 &62.94 &67.23 &79.30\\
                \multicolumn{2}{c|}{BAM (CVPR'2022) \cite{lang2022learning}} &63.18 &\blu{70.77} &\blu{66.14} &\blu{57.53} &\blu{64.41} &\blu{77.26} &67.36 &\blu{73.05} &\blu{70.61} &\blu{64.00} &\blu{68.76} &\blu{81.10}\\
                \multicolumn{2}{c|}{FECANet (TMM'2023) \cite{liu2023fecanet}} &\red{66.50} &68.90 &63.60 &58.30 &64.30 &76.20 &\blu{68.60} &70.80 &66.70 &60.70 &66.70 &77.60\\
           
			\multicolumn{2}{c|}{Ours}    & \blu{64.26} & \red{72.19} &\red{67.65} &\red{57.61} & \red{65.43} & \red{77.81} &\red{69.18} &\red{74.49} &\red{72.56} &\red{64.08} &\red{70.08} &\red{81.95}\\ 
			\hline
			\hline
			\multicolumn{14}{c}{Resnet50-Backbone}  \\ \hline
			\multicolumn{2}{c|}{CANet (ICCV'2019) \cite{canet}}    & 52.50 & 65.90 & 51.30 & 51.90 & 55.40 &66.20 & 55.50 & 67.80 & 51.90 & 53.20 & 57.10 &69.60\\  
			\multicolumn{2}{c|}{PFENet (TPAMI'2020) \cite{PFEnet}}   & 61.70 & 69.50 & 55.40 & 56.30 & 60.80 &73.30 & 63.10 & 70.70 & 55.80 & 57.90 & 61.90 &73.90\\ 
   			\multicolumn{2}{c|}{HSNet (ICCV'2021) \cite{ min2021hypercorrelation}} &64.30 &70.70 &60.30 &60.50 &64.00 &76.70 &70.30 &73.20 &67.40 &67.10 &69.50 &80.60\\ 
            \multicolumn{2}{c|}{NTRENet (CVPR'2022) \cite{liu2022learning}} &65.40 &72.30 &59.40 &59.80 &64.20 &77.00 &66.20 &72.80 &61.70 &62.20 &65.70 &78.40\\ 
            \multicolumn{2}{c|}{DCAMA (ECCV'2022) \cite{shi2022dense}} &67.50 &72.30 &59.60 &59.00 &64.60 &75.70 &70.50 &73.90 &63.70 &65.80 &68.50 &79.50\\
            \multicolumn{2}{c|}{DACM (ECCV'2022) \cite{xiong2022doubly}}  &68.40 &73.10 &63.50 &62.20 &66.80 &78.60 &73.80 &74.70 &70.30 &68.10 &71.70 &81.70 \\
            \multicolumn{2}{c|}{IPMT (NeurIPS'2022) \cite{liu2022intermediate}} &41.40 &45.10 &45.60 &40.00 &43.00&77.10 &43.50 &49.70 &48.70 &47.90 &47.50 &81.40\\
            \multicolumn{2}{c|}{DCP (IJCAI'2022) \cite{beyond_prototype}}&63.81 &70.54 &61.16 &55.69 &62.80 &75.60 &67.19 &73.15 &66.39 &64.48 &67.80 &79.70\\ 
            \multicolumn{2}{c|}{HPA (TPAMI'2022) \cite{HPA}} &67.53 &72.38 &65.21 &56.72 &65.46 &76.40 &71.15 &73.88 &68.83 &63.79 &69.41 &81.10\\
            \multicolumn{2}{c|}{BAM (CVPR'2022) \cite{lang2022learning}} &68.97 &\blu{73.59} &\blu{67.55} &61.13 &\blu{67.81} &\blu{79.71} &70.59 &\blu{75.05} &\blu{70.79} &67.20 &\blu{70.91}&\blu{82.18}\\
            \multicolumn{2}{c|}{FECANet (TMM'2023) \cite{liu2023fecanet}} &\blu{69.20} &72.30 &62.40 &\red{65.70} &67.40 &78.70 &\red{72.90} &74.00 &65.20 &\blu{67.80} &70.00 &80.70\\
			\multicolumn{2}{c|}{Ours}    &\red{69.75} &\red{74.89} &\red{68.91} &\blu{61.55} &\red{68.78} &\red{79.96} &\blu{71.88} &\red{76.63}&\red{71.75} &\red{68.29} &\red{72.14} &\red{83.24}\\ 
			\hline
	\end{tabular}}
	
	\label{pascal}	
 \vspace{-0.2cm}
\end{table*}

\begin{table*}[t!]
	\centering
	\scriptsize
	\footnotesize
	\renewcommand{\arraystretch}{0.9}
	\renewcommand{\tabcolsep}{2.8mm}		
	\caption{ mIoU performance on four folds of COCO-$20^\textit{i}$.\red{Red} and \blu{blue} are leveraged to indicate the two best scores.}
	\scalebox{0.80}{
		\begin{tabular}{cc|cccccc|cccccc}
			\hline
			\multicolumn{2}{c|}{\multirow{2}{*}{Method}}&  \multicolumn{6}{c|}{1-shot}&\multicolumn{6}{c}{5-shot} \\
			
			\multicolumn{2}{c|}{}     
			& Fold-0  & Fold-1 & Fold-2 & Fold-3 & Mean &FB-IoU & Fold-0  & Fold-1 & Fold-2 & Fold-3 & Mean &FB-IoU\\ 
			\hline \hline
			\multicolumn{14}{c}{VGG16-Backbone}  \\ \hline  
			\multicolumn{2}{c|}{FWB  (ICCV'2019) \cite{zhang2022feature}}    & 18.35 &16.72 &19.59 &25.43 &20.02 & - & 20.94 &19.24 &21.94 &28.39 &22.63 & - \\ 
			\multicolumn{2}{c|}{PFENet (TPAMI'2020) \cite{PFEnet}}&35.40 &38.10 &36.80 &34.70 &36.30 &63.30 &38.20 &42.50 &41.80 &38.90 &40.40 &65.00\\ 
                \multicolumn{2}{c|}{HPA (TPAMI'2022) \cite{HPA}} &\blu{40.00} &44.19 &42.16 &41.96 &42.08 &\blu{67.23} &46.22 &51.36 &48.61 &46.30 &48.12 &\blu{72.19}\\
                \multicolumn{2}{c|}{BAM (CVPR'2022) \cite{lang2022learning}}&38.96 &\blu{47.04} &\blu{46.41} &\blu{41.57} &\blu{43.50} &- &\blu{47.02} &\blu{52.62} &\blu{48.59} &\blu{49.11} &\blu{49.34} &-\\
                \multicolumn{2}{c|}{FECANet (TMM'2023) \cite{liu2023fecanet}} &34.10 &37.50 &35.80 &34.10 &35.40 &65.50 &39.70 &43.60 &42.90 &39.70 &41.50 &67.70\\
           
			\multicolumn{2}{c|}{Ours}    &\red{40.71} &\red{49.98} &\red{48.67} &\red{43.78} &\red{45.79} & \red{72.27} &\red{48.70} & \red{56.76} &\red{54.26} &\red{49.14}&\red{52.22} &\red{75.87}\\ 
			\hline
			\hline
			\multicolumn{14}{c}{Resnet50-Backbone}  \\ \hline
			\multicolumn{2}{c|}{PFENet (TPAMI'2020) \cite{PFEnet}} &36.50 &38.60 &35.00 &33.80 &35.80 & - &36.50 &43.30 &38.00 &38.40 &39.00 & -\\ 
   			\multicolumn{2}{c|}{HSNet (ICCV'2021) \cite{ min2021hypercorrelation}}&36.30 &43.10 &38.70 &38.70 &39.20 &68.20 &43.30 &51.30 &48.20 &45.00 &46.90 &70.70\\ 
            \multicolumn{2}{c|}{NTRENet (CVPR'2022) \cite{liu2022learning}} &36.80 &42.60 &39.90 &37.90 &39.30 &68.50 &38.20 &44.10 &40.40 &38.40 &40.30 &69.20\\ 
            \multicolumn{2}{c|}{DCAMA (ECCV'2022) \cite{shi2022dense}} &41.90 &45.10 &44.40 &41.70 &43.30  &69.50 &45.90 &50.50 &50.70 &46.00 &48.30 &71.70\\
            \multicolumn{2}{c|}{DACM (ECCV'2022) \cite{xiong2022doubly}}  &41.20 &45.20 &44.10 &41.30 &43.00 &69.40 &45.20 &52.20 &51.50 &47.70 &49.20 &72.90 \\
            \multicolumn{2}{c|}{IPMT (NeurIPS'2022) \cite{liu2022intermediate}} &41.40 &45.10 &45.60 &40.00 &43.00&- &43.50 &49.70 &48.70 &47.90 &47.50 &-\\
            \multicolumn{2}{c|}{DCP (IJCAI'2022) \cite{beyond_prototype}}&40.89 &43.77 &42.60 &38.29 &41.39 &- & 45.82 &49.66 &43.69 &46.62 &46.48 &-\\ 
            \multicolumn{2}{c|}{HPA (TPAMI'2022) \cite{HPA}} &41.02 &46.85 &44.25 &43.22 &43.84 &68.32 &46.19 &\blu{56.24} &49.20 &50.43 &50.52 &\blu{71.36}\\
            \multicolumn{2}{c|}{BAM (CVPR'2022) \cite{lang2022learning}} &\red{43.41} &\blu{50.59} &\blu{47.49} &\blu{43.42} &\blu{46.23} &- &\red{49.26} &54.20 &\blu{51.63} & \blu{49.55} &\blu{51.16} &-\\
            \multicolumn{2}{c|}{FECANet (TMM'2023) \cite{liu2023fecanet}} &38.50 &44.60 &42.60 &40.70 &41.60 &\blu{69.60} &44.60 &51.50 &48.40 &45.80 &47.60 &71.10\\
			\multicolumn{2}{c|}{Ours}    &\blu{42.48} &\red{53.10} &\red{50.17} &\red{46.36} &\red{48.03} &\red{72.45} &\blu{49.02} &\red{59.08} &\red{53.81} &\red{50.90} &\red{53.20} &\red{73.36}\\ 
			\hline
	\end{tabular}}
	\label{coco}	
 \vspace{0cm}
\end{table*}

\section{Experiments}
\subsection{Datasets and Evaluation Metrics}
We follow previous methods \cite{beyond_prototype, liu2022intermediate,liu2022learning,lang2022learning}  to evaluate our methods in two  standard FFS datasets, i.e.,  PASCAL-$5^\textit{i}$ ~\cite{OSLSM}  and COCO-$20^\textit{i}$ ~\cite{nguyen2019feature} datasets. The combination of the PASCAL VOC 2012 dataset ~\cite{pascal} and the extended SDS dataset ~\cite{SDS} 
 constitutes the PASCAL-$5^\textit{i}$. Normally, the 20 classes in PASCAL-$5^\textit{i}$ would be evenly divided into 4 folds, namely $i \in \left\{ {0,1,2,3} \right\}$. 3 folds would be randomly chosen for training and the rest fold would be used for testing. The overall process would be conducted in a cross-validation manner.  Similarly, COCO-$20^\textit{i}$ is constructed from the MS COCO ~\cite{coco} and the 80 categories of COCO-$20^\textit{i}$ are also evenly divided into 4 folds. Thus, each fold of COCO-$20^\textit{i}$ has 20 classes. For the two standard datasets, we follow the evaluation method in previous algorithms\cite{beyond_prototype, liu2022intermediate,liu2022learning,lang2022learning} to randomly sample 1000 query-support pairs for validation.  Previous methods mainly leverage the  mean intersection-over-union (mIoU) as the evaluation metric, we also adopt this evaluation metric in our paper. Apart from this, we also utilize the foreground-background IoU (FB-IoU) as an additional evaluation metric for better performance analysis.
\subsection{Implementation Details}
Following previous works, we leverage  VGG16 ~\cite{simonyan2014very}, and Resnet50 ~\cite{he2016deep} as the backbones to construct our network for fair comparisons.  The proposed methods are implemented using Pytorch ~\cite{paszke2019pytorch}. All experiments are conducted on the Nvidia Tesla A100 GPUs. Since our method leverages the BAM ~\cite{lang2022learning} as the baseline. The training process is split into two training phases. In the base training phase,  the standard supervised learning paradigm has been leveraged to train the base learner (PSPNet) for base categories of each fold, namely 16/61 classes(including background) for PASCAL-$5^\textit{i}$ and COCO-$20^\textit{i}$. The base learner is trained on PASCAL-$5^\textit{i}$  for 100 epochs and COCO-$20^\textit{i}$ for 20 epochs. We adopt  SGD as the optimizer with a learning rate of 2.5e-3. The batch size of the training base learner is set to 12. For the meta-training phase, we utilize the same training parameters of BAM. First, the feature extractor (backbone) is not optimized with initial pre-trained weights to improve generalization. Then, the rest network is trained with SGD. The learning rate and batch size are set to 5e-2 and 8 for both two datasets. Following the strategy in BAM [19], multiple prototypes are first combined into one prototype in a weighted  fusion manner. Then, Reflection Invariance Prototype Creation is performed. The training process lasts 200 epochs and 50 epochs respectively for PASCAL-$5^\textit{i}$ and COCO-$20^\textit{i}$.

\subsection{ Comparison with State-of-the-art Methods}
\textbf{Quantitative Results.} Table  \ref{pascal} reports the performance comparisons in PASCAL-$5^\textit{i}$ dataset. Clearly, our proposed method could achieve the best performance. More specifically, with the VGG16 backbone,   our method acquires 65.43 and 70.08 averaged mIoU scores respectively under 1-shot and 5-shot settings. Compared with the previous best method, our model respectively brings 1.02\% and 1.32\% mIOU improvements for 1-shot and 5-shot settings. Simultaneously, our method could also bring improvements in FBI-Iou scores. With the Resnet50 backbone, our method obtains 68.78 and 72.14 averaged mIoU scores respectively under 1-shot and 5-shot settings, which surpass state-of-the-art results by 0.97\% and 1.23\%. Meanwhile, though the FB-IoU of previous works are very high, our method could also bring some improvements.

Table \ref{coco} illustrates the performance comparisons in COCO-$20^\textit{i}$ dataset. As shown in Table \ref{coco}, though COCO-$20^\textit{i}$ dataset is more challenging, our method could still achieve great performance. Particularly, with the VGG16 backbone,  our model could obtain 45.79 and 52.22 averaged mIOU scores under 1-shot and 5-shot settings. Clearly, our model brings 2.29\%  and 2.88 \% mIOU improvements. Moreover, our method also brings high FB-IoU performance improvements compared with the previous best method, namely 5.04 \% and 3.69 \% respectively for 1-shot and 5-shot settings. With the Resnet50 backbone, our model obtains 48.03 and 53.20 averaged mIOU scores under 1-shot and 5-shot settings, which achieves 1.8\%  and 2.04 \% mIOU improvements. Compared with previous best results, the large improvement (2.85 \% for 1-shot, 2.00\% for 5-shot) of FB-IoU could also demonstrate the superiority of our proposed FSS method.

\begin{table}[t]
	\centering
	\scriptsize
	\footnotesize
	\renewcommand{\arraystretch}{1.0}
	\renewcommand{\tabcolsep}{3.0mm}
	\caption{Ablation study of support and query reflection invariance. SFI denotes support reflection invariance and QFI denotes query reflection invariance. }
	\scalebox{1.0}{
		\begin{tabular}{cc|ccccc}
			\hline
			\multicolumn{1}{c}{\multirow{1}{*}{SFI}}&  \multicolumn{1}{c|}{QFI} &\multicolumn{1}{c}{Fold-0}  &\multicolumn{1}{c}{Fold-1}  &\multicolumn{1}{c}{Fold-2} &\multicolumn{1}{c}{Fold-3}&\multicolumn{1}{c}{Mean} \\
			\hline
			\multicolumn{1}{c}{} &    \multicolumn{1}{c|}{}  &68.97 &73.59 &67.55 &61.13 &67.81 \\ 
			\multicolumn{1}{c}{\checkmark} & \multicolumn{1}{c|}{} &69.87 &75.11 &67.42 &61.81  &68.55 \\
   			\multicolumn{1}{c}{} & \multicolumn{1}{c|}{\checkmark} &69.37 &74.51 &67.64 &61.01  &68.13 \\
			\multicolumn{1}{c}{\checkmark} &  \multicolumn{1}{c|}{\checkmark} &\textbf{69.75} &\textbf{74.89} &\textbf{68.91} &\textbf{61.55} &\textbf{68.78} \\  \hline		
	\end{tabular}}
	\label{table3}
 \vspace{-0.4cm}
\end{table}

\begin{figure}[t]
	\begin{center}
		\includegraphics[width=1.0\linewidth]{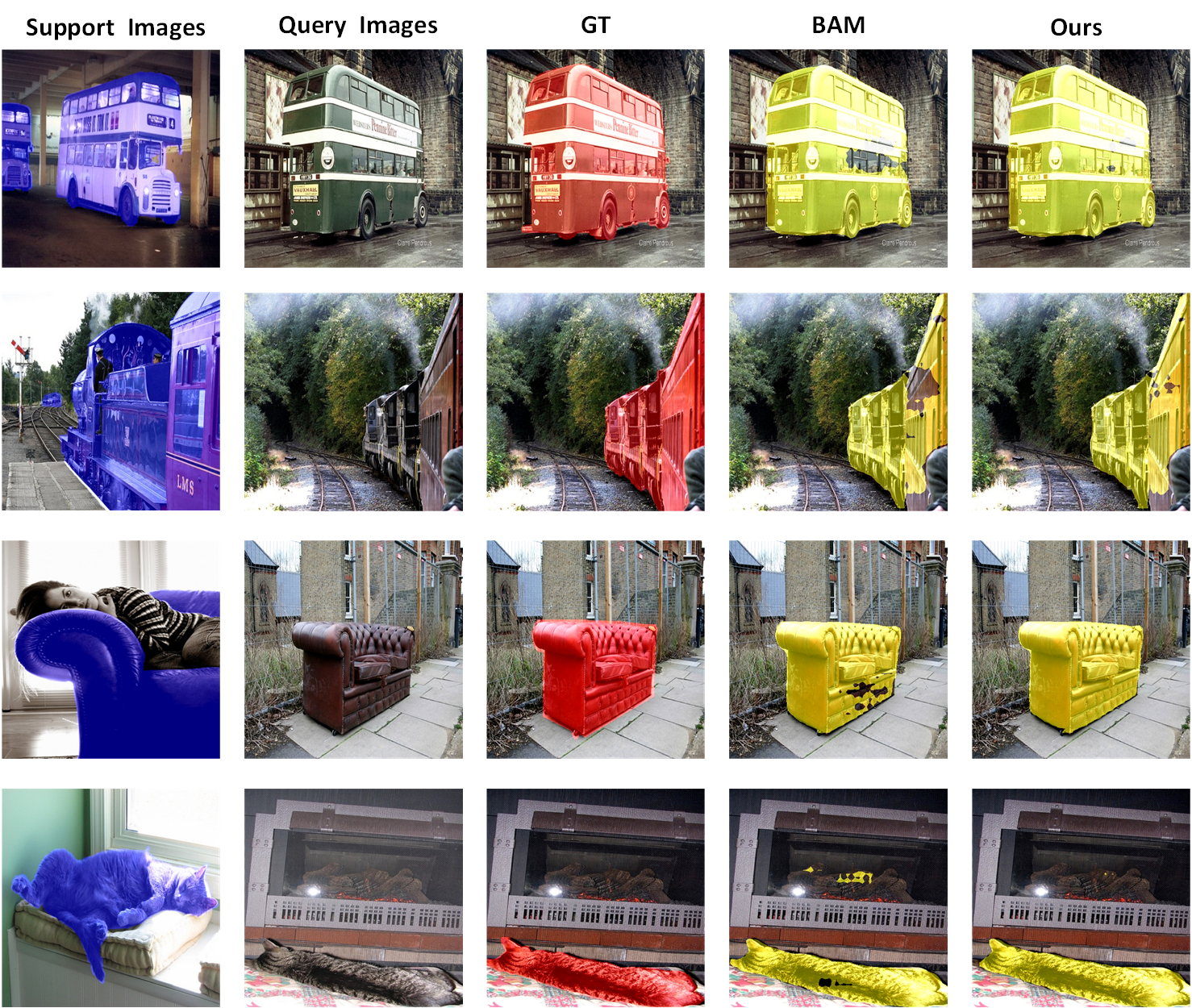}
	\end{center}
	\caption{Qualitative results of our proposed model. From left to right: support images, query images, ground truths, predictions of BAM, and predictions of our method. }
	\label{fig5}
  \vspace{-0.5cm}
\end{figure}

\textbf{Qualitative Results.} We show the qualitative results of our model in Figure \ref{fig5}. We utilize the visualization results of BAM as comparisons to provide better analysis. As shown in Figure \ref{fig5},  support images, query images, ground truth, the result of BAM, and the result of our model are illustrated sequentially in each row. We could see that our proposed model could have better performance. For instance, more foreground parts could be precisely parsed. Apart from this, we could also clearly see that some false activating regions could be eliminated.

\subsection{Ablation Study}
We set a series of ablation experiments to demonstrate the effectiveness of our model. The experiments are conducted in the PASCAL-$5^\textit{i}$ dataset with Resnet50 backbone under the 1-shot setting.

\begin{table}[t]
	\centering
	\scriptsize
	\footnotesize
	\renewcommand{\arraystretch}{1.0}
	\renewcommand{\tabcolsep}{3.0mm}
	\caption{ Ablation study of reflection. Aug denotes flipping data augmentation strategy}
	\scalebox{1.0}{
		\begin{tabular}{c|ccccc}
			\hline
			\multicolumn{1}{c|}{Folds}  &\multicolumn{1}{c}{Fold-0}  &\multicolumn{1}{c}{Fold-1}  &\multicolumn{1}{c}{Fold-2} &\multicolumn{1}{c}{Fold-3} &\multicolumn{1}{c}{Mean} \\
			\hline
   		\multicolumn{1}{c|}{Baseline} &68.01 & 73.26 &67.52 &61.29 &67.52 \\
            \multicolumn{1}{c|}{Baseline+Aug}  &68.97 &73.59 &67.55 &61.13 &67.81 \\
			\multicolumn{1}{c|}{Ours} &\textbf{69.75} &\textbf{74.89} &\textbf{68.91} &\textbf{61.55} &\textbf{68.78}  \\
			\hline	
	\end{tabular} }
	\label{table4}
 \vspace{-0.4cm}
\end{table}

\textbf{Ablation Study on Variance.} Since the main idea of our work is to mine the semantic reflection invariance, we need to analyze the effectiveness of the leveraged support and query reflection invariance. In our model, we leverage the reflection invariance prototypes to mine the support reflection invariance (SRI). The RIPMG and RISP are leveraged to fuse the information obtained from the query reflection invariance (QRI). Thus, Removing SRI refers to removing reflection invariance prototypes. Removing QRI means removing the RIPMG and RISP modules. The experimental results of the two variances are shown in Table \ref{table3}. As can be seen,  mining a single SRI or QRI could also lead to some performance improvement. The combination of SRI and QRI could result in obvious performance improvements. RISP fuses same-supervised predictions in varying views to obtain consistent predictions. To analyze the effectiveness of RISP, independent experiments are performed and the results are shown in Table~\ref{table5}. Clearly, a single RISP could also bring performance enhancement.

\begin{table}[t]

	\centering
	\setlength{\tabcolsep}{2.5mm}
	\caption{The ablation study of RISP. 
	}
 
	\resizebox{1.0\columnwidth}{!}{%
\begin{tabular}{c|ccccc}
\hline
Folds         & Fold-0 & Fold-1 & Fold-2 & Fold-3 & Mean  \\ \hline
Baseline      & 68.97  & 73.59  & 67.55  & 61.13  & 67.81 \\
Baseline+RISP  & \textbf{69.25} & \textbf{74.38} & \textbf{67.60} & \textbf{61.16} & \textbf{68.10} \\ \hline
\end{tabular}
	}
	\label{table5}%
\end{table}

\begin{table}[t]
	\scriptsize
	\footnotesize
	\renewcommand{\arraystretch}{1.0}
	\renewcommand{\tabcolsep}{3.0mm}
	\caption{Ablation study of data augmentation. \textbf{RE}: Random Exposure $\left[0.25,4\right]$, \textbf{RS}: Random Scale $\left[0.5,1.5\right]$, \textbf{RR}: Random Rotation $\left[0^{\circ},360^{\circ}\right]$ 
	}
	\resizebox{1.0\columnwidth}{!}{%
\begin{tabular}{c|ccccc}
\hline
Folds         & Fold-0 & Fold-1 & Fold-2 & Fold-3 & Mean  \\ \hline
Baseline      & 68.97 &73.59 &67.55 &61.13 &67.81\\
RE   & 67.07 &71.49 &66.55 &60.13 &66.31\\
RS   & 68.86 &73.69 &67.85 &61.33 &67.93\\
RR   & 68.67 &72.99 &66.95 &61.11 &67.43\\
Ours & \textbf{69.75} &\textbf{74.89} &\textbf{68.91} &\textbf{61.55} &\textbf{68.78} \\ \hline
\end{tabular}
	}
	\label{table6}%
\end{table}

\begin{table}[t]
	\caption{The study of multiple support images with diverse augmentations. 
	}
	\scriptsize
	\footnotesize
	\renewcommand{\arraystretch}{1.0}
	\renewcommand{\tabcolsep}{3.0mm}
	\resizebox{1.0\columnwidth}{!}{%
\begin{tabular}{c|ccccc}
\hline
Folds         & Fold-0 & Fold-1 & Fold-2 & Fold-3 & Mean  \\ \hline
Ours+RS      & 69.20  & 74.43  & 67.55  & 61.04  & 68.06 \\
Ours  & \textbf{69.25} & \textbf{74.38} & \textbf{67.60} & \textbf{61.16} & \textbf{68.10} \\ \hline
\end{tabular}
	}
	\label{table7}%
\end{table}

 \begin{figure}[t]
	\begin{center}
		\includegraphics[width=1.0\linewidth]{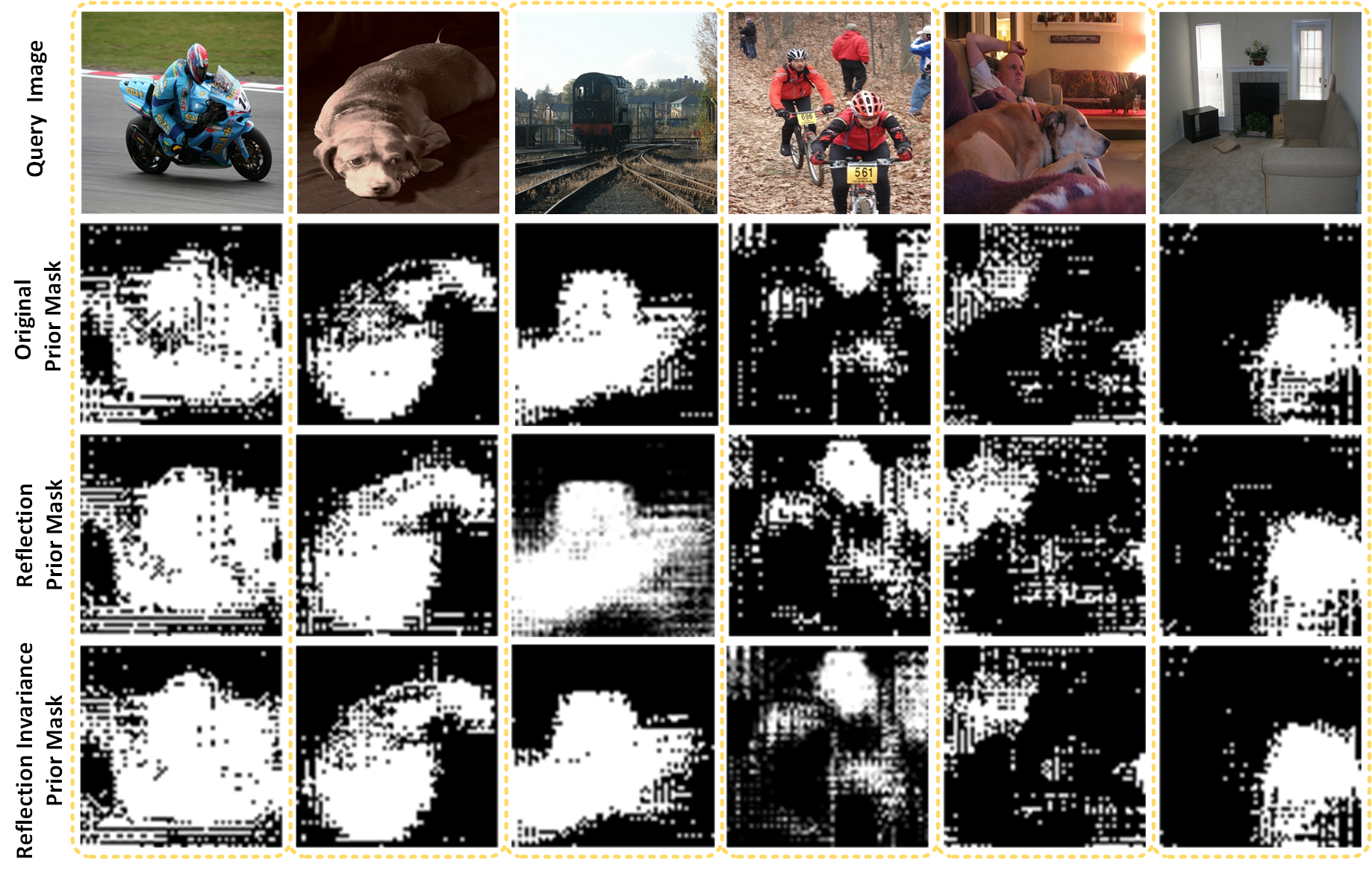}
	\end{center}
	\caption{Visualization of reflection invariance prior masks.  From top to down: query images, original prior masks, reflection prior masks, and reflection invariance prior masks. }
	\label{fig6}
  \vspace{-0.5cm}
\end{figure}

 \begin{figure}[t]
	\begin{center}
		\includegraphics[width=1.0\linewidth]{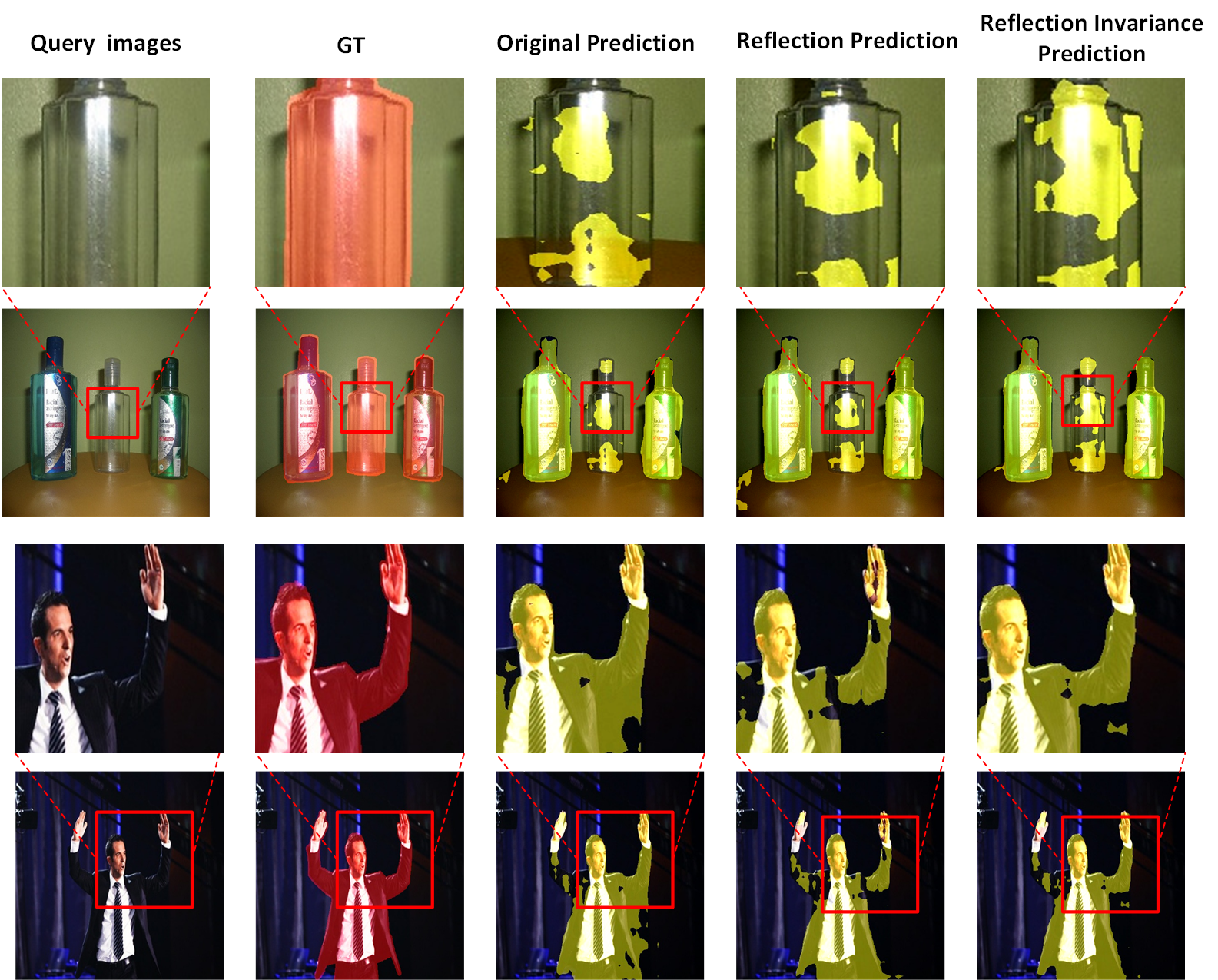}
	\end{center}
	\caption{Visualization of reflection invariance predictions.  From left to right: query images, ground truths, original predictions, reflection predictions, and reflection invariance predictions. }
	\label{fig7}
  \vspace{-0.5cm}
\end{figure}
\textbf{Ablation Study on Reflection.}  Many deep learning-based networks would adopt random flipping as a data augmentation strategy. Different from this augmentation strategy that only works on input, our methods utilized elaborately designed architectures to simultaneously explore support and query reflection invariance. To further distinguish our model from this augmentation strategy, we conducted some experiments and the experimental results are shown in Table \ref{table4}. As we could see, different from the traditional data augmentation strategy, our proposed model could obtain more obvious performance improvements. Other operations( like  Random Exposure (RE),  Random Scale (RS), and Random Rotation (RR)) might also be helpful to mine the corresponding invariance. Thus, we  replace the change views with various operations, and set a serial of experiments to compare performance. As shown in Table~\ref{table6}.  Only RS obtain tiny improvements, others get degradation. Thus reflection is the most critical and beneficial factor to mine the invariance.  Following the strategy in BAM [19], multiple prototypes are first combined into one prototype in a weighted  fusion manner. Then, Reflection Invariance Prototype Creation is performed. Multiple support images could be supplied with diverse augmentations to provide different views. However, since only RS brings tiny enhancement, we directly fuse the results of RS and ours to compare performance (shown in Tab.~\ref{table7}). The similar performance implies our method is enough to mine invariance

\textbf{Study on Reflection Invarinace Prior Mask.} To better analyze the function of reflection invariance prior mask. Some visualization results are shown in Figure \ref{fig6}. We could see original and reflection prior masks have different activation and inhibition areas. Simultaneously, the same activation or inhibition areas have different activation values (brightness).  By efficiently fusing the two prior masks, the generated reflection invariance prior masks clearly have better support guidance.

\textbf{Study on Reflection Invarinace Semantic Prediction.} Aiming at directly analyzing the performance of the reflection invariance prediction, some visualization results of reflection invariance predictions are shown in Figure \ref{fig7}. Obviously, original predictions and reflection predictions have different focusing areas, which corresponds to the fact that different perspectives always lead to different results. By rationally fusing the two predictions, the generated reflection invariance predictions have a better performance than the original predictions and reflection predictions. This phenomenon sufficiently proves the effectiveness of our proposed method.

\section{Conclusion}
In this work, we propose a new multi-view matching framework to mine the semantic reflection invariance for few-shot semantic segmentation. Particularly, through mining and utilizing the support reflection invariance, the generated reflection invariance prototypes have stronger category representative ability. By excavating and leveraging the query reflection invariance, predictions from different perspectives are fused to provide better reflection invariance  predictions. Surprisingly,  even with such a straightforward framework, our method outperforms previous methods in terms of segmentation performance. We hope our work could inspire future research to concentrate more on mining the inner characteristics, e.g. semantic reflection invariance, in FSS.


{\small
\bibliographystyle{ieee_fullname}
\bibliography{egpaper_for_review}
}

\end{document}